\documentclass[letterpaper]{article} 
\usepackage{aaai2027}  
\usepackage[hyphens]{url}  
\usepackage{graphicx} 
\usepackage{natbib}  
\usepackage{caption} 
\usepackage{algorithm}
\usepackage{algorithmic}
\usepackage{amssymb}
\usepackage{multirow}
\usepackage{subfig}
\usepackage[table]{xcolor}
\usepackage{newfloat}
\usepackage{listings}
\DeclareCaptionStyle{ruled}{labelfont=normalfont,labelsep=colon,strut=off} 
\floatstyle{ruled}
\newfloat{listing}{tb}{lst}{}
\floatname{listing}{Listing}

\usepackage{booktabs}
\usepackage{amsmath}

\usepackage{adjustbox}
\usepackage{ragged2e}
\newcommand{\textfontsize}{6}
\newcommand{\textbaselineskip}{8}
\newcommand{\headfontsize}{6}
\newcommand{\headbaselineskip}{9}
\newcommand{\rowspacing}{-3pt}
\newcommand{\headrowspacing}{-7pt}
\newcommand{\colspacing}{0.5pt}
\newcommand{\textcolgap}{3pt}
\newcommand{\arraystretchval}{1.1}

\newcommand{\headcell}[1]{\parbox[c][0.25in][c]{\linewidth}{\centering\fontsize{\headfontsize}{\headbaselineskip}\selectfont #1}}
\newcommand{\textcell}[1]{\parbox[b][\imgsize][c]{\linewidth}{\RaggedRight\fontsize{\textfontsize}{\textbaselineskip}\selectfont #1}}
\newcommand{\imgsubfig}[2][1]{%
  \includegraphics[page=#1,width=\imgsize,height=\imgsize]{#2}%
}

\title{Exploring the Potential of Contrastive Language-Image Pre-training for Multi-Source Remote Sensing Data}
\author{
    Xiangyang Miao\textsuperscript{\rm 1} \equalcontrib,
    Kelu Yao\textsuperscript{\rm 2} \equalcontrib,
    Yekai Huang\textsuperscript{\rm 1}, 
    Xiaogang Xu \textsuperscript{\rm 1},
    Junxiao Xue\textsuperscript{\rm 2},
    Minjun Shen\textsuperscript{\rm 2},
    Chenghui Lv\textsuperscript{\rm 1},
    Shanji Liu\textsuperscript{\rm 1},
    Yaying Chen\textsuperscript{\rm 1},
    Chao Li\textsuperscript{\rm 2}
}
\affiliations{
    \textsuperscript{\rm 1}School of Computer Science and Technology, Zhejiang University\\
    \textsuperscript{\rm 2}Space-based Computing System Research Center, Zhejiang Lab
}

\begin{document}

\maketitle

\begin{abstract}
Contrastive language-image learning (CLIP) has become a key paradigm for remote sensing vision-language understanding. However, existing remote sensing contrastive learning methods are mostly built on RGB-oriented CLIP architectures, making it difficult to exploit heterogeneous sensors such as SAR, multi-spectral imaging (MSI), and hyperspectral imaging (HSI). To address this limitation, we propose OmniRSCLIP, an end-to-end contrastive learning framework that supports multi-source sensor inputs for remote sensing vision-language modeling. The key idea is to extend CLIP beyond its fixed RGB input interface without breaking the pretrained visual knowledge. To this end, OmniRSCLIP introduces Spectral-Spatial Basis Decomposition (SSBD), which formulates arbitrary-channel adaptation as a basis recomposition problem: pretrained CLIP patch embeddings provide transferable spatial bases, while wavelength-conditioned coefficients span sensor-specific embedding kernels within a constrained visual prior space. This design avoids forcing heterogeneous sensors into a fixed-channel input space, while aligning them in a unified image-text semantic space. We further introduce a spectral-context-aware mask-based contrastive learning scheme to suppress modality-specific redundant features and enhance fine-grained image-text alignment. Finally, to support multi-modal training, we construct OmniRS5M, the first large-scale remote sensing image-text corpus covering RGB, SAR, MSI, and HSI. Experiments on retrieval, zero-shot classification, and semantic localization show that OmniRSCLIP preserves strong RGB-domain performance while effectively extending CLIP to heterogeneous remote sensing modalities.

\end{abstract}


\section{Introduction}

\begin{figure}[t]
  \centering
   \includegraphics[page=1,width=1\linewidth]{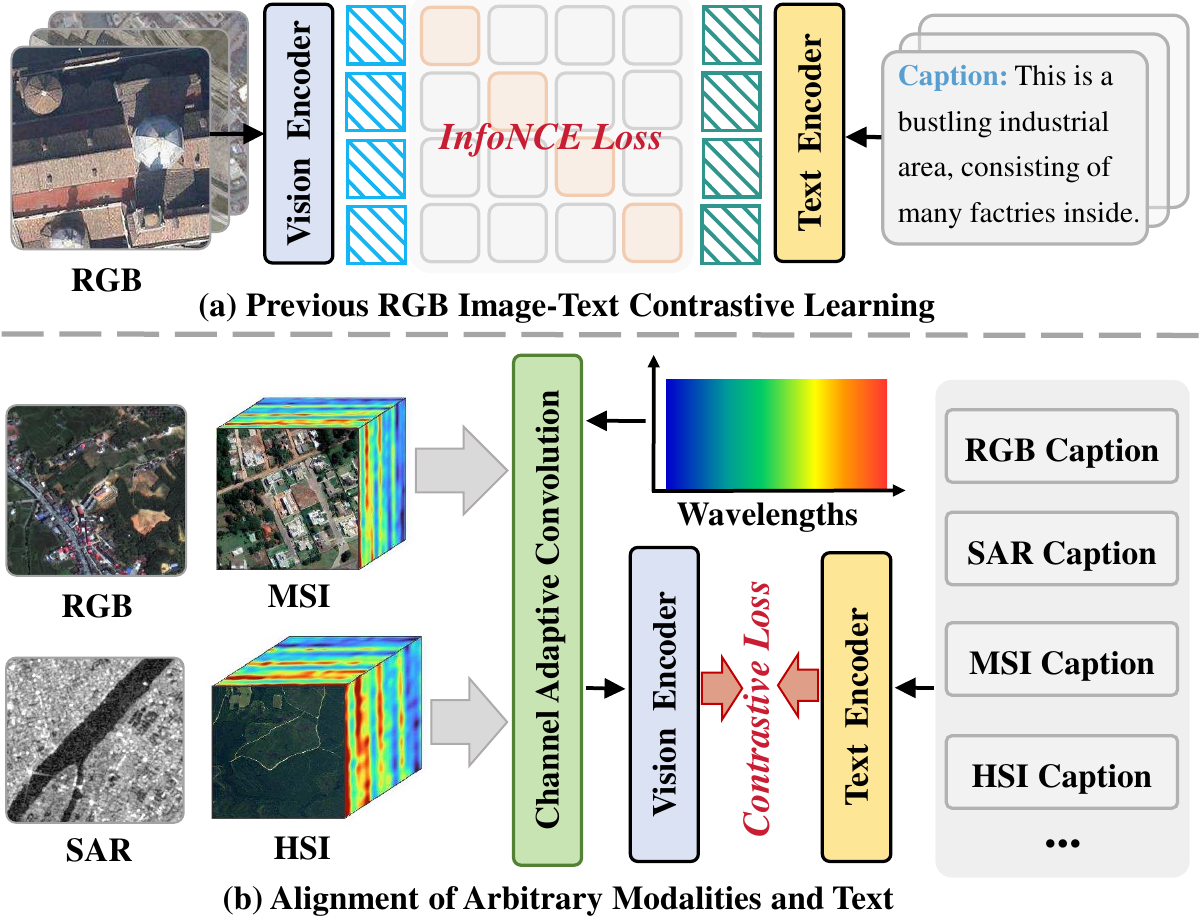}
   \caption{\textbf{Comparison of remote sensing image-text alignment paradigms.} (a) Existing methods are mainly built for RGB imagery and are difficult to extend to heterogeneous sensor inputs. (b) OmniRSCLIP uses SSBD as a modality-adaptive visual embedding interface, enabling a single model to process RGB, SAR, MSI, and HSI data within a unified image-text alignment framework.}
   \label{fig:intr}
\end{figure}

Language-image contrastive learning, exemplified by CLIP~\cite{clip}, has achieved remarkable progress in cross-modal semantic alignment and transferable representation learning, supporting tasks such as image-text retrieval~\cite{huang2026llm2clip,qin2025clip,siglip2}, open-vocabulary segmentation~\cite{liu2025unveiling,li2026exploring} and anomaly detection~\cite{gao2026adaptclip,ma2025aa}. This contrastive pre-training paradigm has also been introduced into the remote sensing community, offering a promising route toward general-purpose earth observation representations. However, unlike natural images that are typically represented as RGB inputs, remote sensing data are collected from diverse sensors~\cite{wolf2026brewing,yang2026sar}, including optical, SAR, MSI, and HSI, which differ substantially in channel numbers, spectral responses, and imaging mechanisms. Most existing remote sensing contrastive learning frameworks~\cite{dgtrsclip,geoalignclip} remain centered on RGB-oriented CLIP architectures or optical imagery, making it difficult to exploit heterogeneous sensor inputs in a unified manner. Therefore, developing an end-to-end remote sensing contrastive learning framework that can accommodate arbitrary-channel modalities while retaining the transferable knowledge of CLIP remains an important challenge.

This challenge first raises a fundamental model-design question: how can an RGB-oriented CLIP model be extended to heterogeneous remote sensing sensors without sacrificing the pretrained visual knowledge that makes contrastive transfer effective? A straightforward strategy is to transform non-RGB data into a three-channel RGB-compatible format, for example, by projecting MSI and HSI data with PCA~\cite{gong2025bridging,zhao2025dynamic}. Although this enables direct use of the original CLIP visual encoder, rich spectral observations are forced into a fixed input format, which can discard sensor-specific information. For example, HSI captures material-specific absorption signatures across hundreds of contiguous narrow bands, whereas projecting them onto only three principal components may discard subtle but discriminative spectral differences. Recent arbitrary-channel embedding methods~\cite{msclip,dofa} offer a more flexible direction, but they often rely on reinitialized or independently generated patch embedding layers. During end-to-end optimization, this may break the compatibility between the embedding head and the pretrained CLIP visual encoder, forcing the model to relearn low-level spectral-spatial projections and weakening the inherited RGB-domain prior. \textit{Therefore, the key issue is not merely to accept more input channels, but to build an adaptation mechanism that remains anchored to the pretrained CLIP embedding space while modeling sensor-specific spectral responses.} Meanwhile, data supervision also limits multi-modal remote sensing contrastive learning: existing optical image-text datasets~\cite{rsicd,rsitmd,ucmcaption} mainly contain coarse scene-level captions, while paired language supervision for SAR, MSI, and HSI remains scarce.

To address these challenges, we propose OmniRSCLIP, the first end-to-end contrastive learning framework that supports multi-source sensor inputs for remote sensing vision-language modeling, as shown in Fig.~\ref{fig:intr}. The core of OmniRSCLIP is the SSBD module, which serves as a modality-adaptive visual embedding interface between heterogeneous sensor inputs and the CLIP visual backbone. Rather than treating arbitrary-channel adaptation as an unconstrained kernel prediction problem, SSBD formulates it as a basis recomposition problem. Specifically, the pretrained CLIP patch embedding provides transferable spatial bases that retain generic local visual primitives, while wavelength-conditioned coefficients recompose channel-specific spectral-spatial kernels within this constrained visual prior space. This formulation reflects a natural property of multi-source remote sensing observations: different spectral bands and sensors often reflect similar spatial structures but require different spectral responses~\cite{hyperfree,spectralgpt}. Consequently, OmniRSCLIP can align RGB, SAR, MSI, and HSI observations within a unified image-text semantic space without forcing heterogeneous data into a three-channel representation or relearning the input layer from scratch. Furthermore, although SSBD preserves sensor-specific information from heterogeneous inputs, not all resulting visual features contribute equally to alignment with rich textual descriptions. We therefore develop a spectral-context-aware mask-based contrastive learning scheme, emphasizing the most text-relevant visual features to further refine cross-modal alignment.

Finally, to support large-scale multi-modal training, we construct OmniRS5M, the first large-scale remote sensing image-text corpus covering RGB, SAR, MSI, and HSI modalities. OmniRS5M is built through a hybrid data collection and modality-aware caption distillation pipeline. Building on existing reliable image-text datasets, we further expand language supervision by converting visual-only RGB, SAR, MSI, and HSI datasets using modality-specific prompts and advanced VLMs. The resulting global scene descriptions and local details are further recombined into short and long caption candidates, providing multi-granularity supervision for contrastive pre-training. After filtering and manual quality inspection, OmniRS5M contains 4.67 million images and 73.90 million text candidates, including both short and long captions, offering the data foundation for exploring contrastive language-image pre-training on multi-source remote sensing data.

The main contributions of this paper are summarized as follows:
\begin{itemize}
  \item We propose OmniRSCLIP, an end-to-end remote sensing vision-language framework for unified modeling of heterogeneous sensor inputs. To bridge arbitrary-channel imagery with the CLIP visual backbone, we introduce SSBD as a modality-adaptive visual embedding interface, which generates wavelength-aware spectral-spatial kernels while inheriting the RGB-domain visual prior of CLIP.
  \item We construct OmniRS5M, the first large-scale multi-modal remote sensing image-text corpus covering RGB, SAR, MSI, and HSI modalities. With modality-aware caption distillation and global-local text synthesis, OmniRS5M provides diverse multi-granularity supervision for learning unified remote sensing vision-language representations.
  \item Extensive experiments on short-text, long-text, and multi-modal retrieval benchmarks validate the effectiveness of OmniRSCLIP. The results demonstrate its advantages over existing remote sensing vision-language models and alternative arbitrary-channel adaptation strategies across heterogeneous remote sensing modalities.
\end{itemize}

\section{Related Work}
\subsection{Contrastive Language-Image Pre-training}
CLIP~\cite{clip} learns a shared vision-language embedding space through large-scale image-text contrastive pre-training, providing a solid foundation for open-vocabulary recognition and image-text retrieval. Recent studies in the natural image domain further show that extending the text context and strengthening fine-grained semantic alignment can substantially improve CLIP's ability to understand complex scenes~\cite{longclip,finelip,fgclip,himoclip}. These advances provide an important basis for transferring CLIP-style contrastive learning to remote sensing. In the remote sensing community, early studies first fine-tune CLIP with large-scale short-caption satellite image-text datasets, including RemoteCLIP~\cite{remoteclip}, GeoRSCLIP~\cite{georsclip}, and SkyScript~\cite{skyscript}. Later works gradually move toward richer semantic supervision: DGTRS-CLIP~\cite{dgtrsclip} introduces dual-granularity semantic alignment to improve the understanding of complex remote sensing scenes, while GeoAlignCLIP~\cite{geoalignclip} explicitly enforces global-local multi-view consistency. However, unlike natural images, remote sensing data are acquired by diverse sensors with different channel numbers, spectral responses, and imaging mechanisms. Existing methods are still primarily designed for RGB imagery. Although several studies have made preliminary attempts to extend contrastive learning to MSI and HSI data, they often require separate visual encoders or modality-specific embedding heads for different sensors, making it difficult to build a unified framework for heterogeneous remote sensing inputs~\cite{timesenclip,msclip}. In contrast, OmniRSCLIP supports multi-source remote sensing inputs within a single contrastive learning framework while preserving the pretrained CLIP visual prior.

\subsection{Channel-Adaptive Embedding}
Earth observation data exhibit highly diverse channel configurations, ranging from single-channel SAR data to HSI with hundreds of narrow bands. To accommodate such variable-channel inputs, recent studies have explored several forms of channel-adaptive visual embedding. HyperFree~\cite{hyperfree} adopts wavelength-aware dictionaries but suffers limited cross-modality transferability due to fixed spectral range constraints. EarthDial~\cite{earthdial} processes arbitrary modalities by iteratively grouping input channels into three-channel subsets, while Any-Optical-Model~\cite{li2026any} applies a shared convolutional kernel across spectral bands; both strategies improve input compatibility but may insufficiently model channel-specific spectral responses. The most related method to ours is DOFA~\cite{dofa}, which dynamically predicts patch-embedding weights from spectral wavelengths. However, generating the full embedding kernel introduces a large prediction space and may reduce compatibility with the pretrained CLIP embedding prior. In contrast, SSBD formulates arbitrary-channel adaptation as compact spectral-spatial basis recomposition, preserving CLIP-inherited spatial priors while predicting only wavelength-conditioned basis coefficients.

\section{Method}
This section details OmniRSCLIP, an end-to-end remote sensing vision-language framework that generalizes CLIP to heterogeneous Earth observation modalities. OmniRSCLIP addresses arbitrary-channel sensor inputs through its core SSBD module, which serves as a modality-adaptive visual embedding interface between multi-source imagery and the CLIP visual backbone. In addition, a spectral-context-aware mask-based contrastive learning scheme is adopted during training to further improve cross-modal alignment.

\begin{figure*}[ht]
\centering
\includegraphics[width=0.98\linewidth]{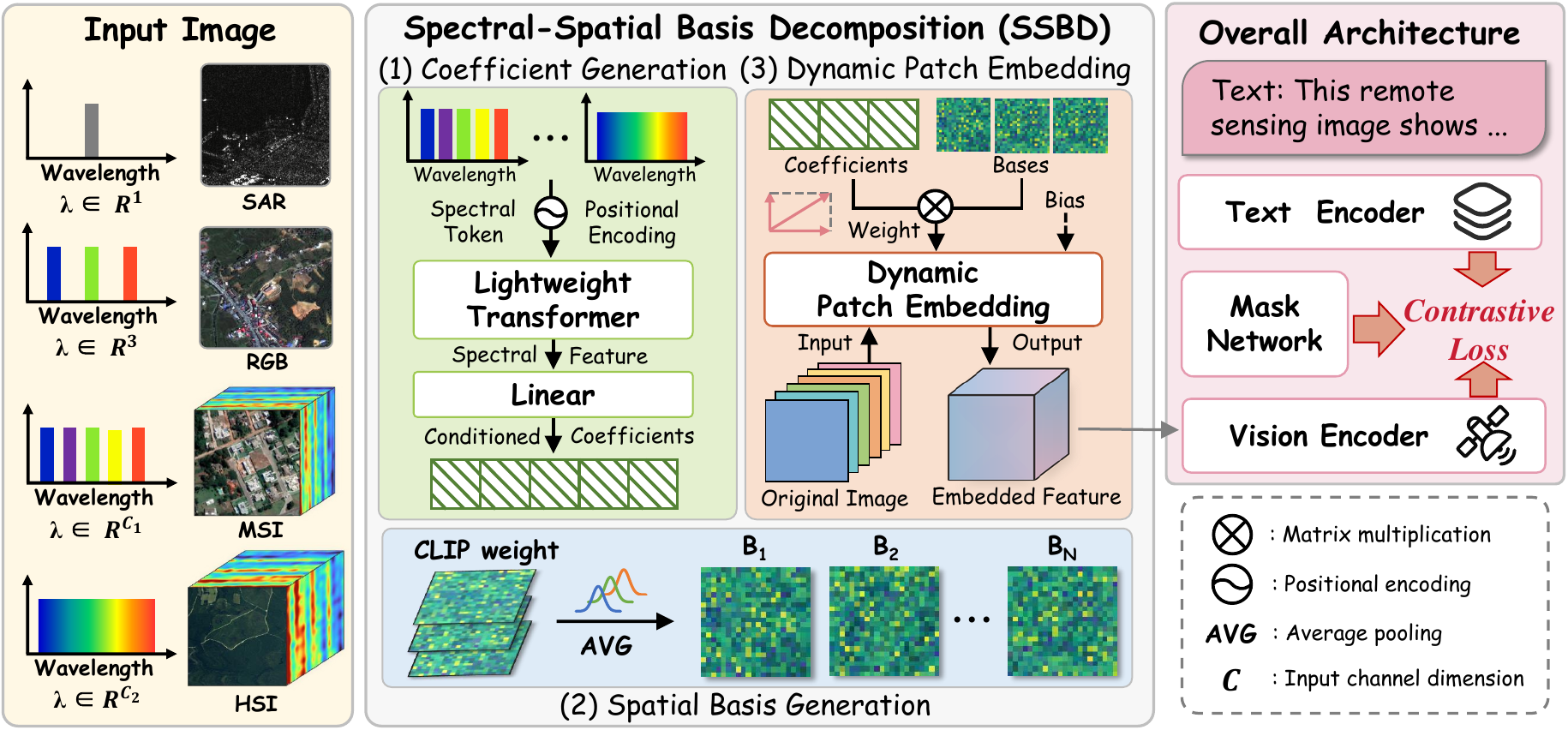}
\caption{\textbf{Overview of the proposed OmniRSCLIP framework.} The key contribution is SSBD, a wavelength-conditioned input interface that dynamically generates embedding weights to adapt multi-source remote sensing images to a shared CLIP visual backbone, followed by feature selection guided by both text and spectral context via a mask network for image-text alignment.}
\label{fig:framework}
\end{figure*}

\subsection{Architecture Overview}
As illustrated in Fig.~\ref{fig:framework}, OmniRSCLIP follows the dual-tower design of CLIP, where visual and text encoders project remote sensing images and language descriptions into a shared contrastive space. To support heterogeneous sensors, we replace the fixed RGB patch embedding with SSBD: given an arbitrary-channel image $\mathbf{X}\in\mathbb{R}^{C\times H\times W}$, where $C$ denotes the number of input channels and $H$ and $W$ denote the image height and width, together with its wavelength sequence $\boldsymbol{\lambda}=\{\lambda_i\}_{i=1}^{C}$, SSBD dynamically generates spectral-spatial kernels to extract visual features compatible with the CLIP visual backbone. Thus, RGB, SAR, MSI, and HSI inputs can share the same visual backbone. For the language branch, we employ KPS~\cite{longclip} to extend the context length from 77 to 248 tokens for long remote sensing descriptions.  Moreover, we extend the mask-based alignment mechanism~\cite{smartclip} by injecting SSBD-derived spectral context into the Mask Network, enabling caption- and sensor-aware feature selection for refined cross-modal alignment.

\subsection{Motivation}

The central difficulty in adapting CLIP to multi-source remote sensing data is the mismatch between the fixed RGB patch embedding and variable-channel sensor observations. The original CLIP patch embedding is parameterized by $\mathbf{W}_{\mathrm{clip}}\in\mathbb{R}^{D\times 3\times K\times K}$, where $D$ denotes the embedding feature dimension and $K$ is the patch size, and is therefore tied to three-channel RGB inputs. A straightforward solution is to project non-RGB data into RGB space, for example by applying PCA to MSI or HSI data before feeding them into CLIP~\cite{gong2025bridging,zhao2025dynamic}. Although this strategy preserves the original CLIP parameters, it compresses high-dimensional spectral observations into three channels and may discard sensor-specific cues. Another representative direction, DOFA~\cite{dofa}, predicts the entire convolutional embedding layer from wavelength information. For an input with $C$ channels, this layer contains $DCK^2$ kernel parameters. This offers arbitrary-channel flexibility, but for high-dimensional sensors, the large parameter-generation space can increase the early optimization burden and cause the generated embeddings to deviate from the pretrained CLIP visual prior.

To build a structured adaptation mechanism, we use the notion of basis representation from linear algebra. For a vector $\mathbf{w}\in\mathbb{R}^{M}$, a set of $M$ linearly independent basis vectors $\{\mathbf{b}_m\}_{m=1}^{M}$ can represent any vector in this space as
\begin{equation}
  \mathbf{w}=\sum_{m=1}^{M}\alpha_m\mathbf{b}_m,
  \quad
  \mathbf{b}_m\in\mathbb{R}^{M},
\end{equation}
where $\alpha_m$ is the coordinate of $\mathbf{w}$ under the $m$-th basis vector. This view separates the representation into two components: a set of basis vectors that define the coordinate system, and a set of coefficients that specify how a particular vector is composed.

SSBD applies this idea to patch embedding kernels, but uses a compact learnable dictionary rather than a complete basis of the full kernel space. For the $i$-th spectral channel, the corresponding channel-wise kernel is $\mathbf{W}_i\in\mathbb{R}^{D\times K\times K}$, which can be flattened into a vector $\mathbf{w}_i\in\mathbb{R}^{M}$ with $M=D K^2$. Instead of predicting all $M$ entries of each $\mathbf{w}_i$ directly, SSBD approximates the channel-wise kernel using $N$ learnable spatial bases $\{\mathbf{b}_n\}_{n=1}^{N}$ with $\mathbf{b}_n \in \mathbb{R}^{M}$:
\begin{equation}
  \mathbf{w}_i
  \approx
  \sum_{n=1}^{N}\alpha_{i,n}(\boldsymbol{\lambda})\mathbf{b}_n,
  \quad
  N \ll M,
\label{eq:vectorized_basis}
\end{equation}
where $\boldsymbol{\lambda}$ is the central wavelength of the sampling sensor channel; $\alpha_{i,n}(\cdot)$ represents a wavelength-conditioned coefficient function that assigns the $n$-th spatial basis to the $i$-th spectral channel according to the full spectral configuration of the input. After reshaping $\mathbf{w}_i$ back to $\mathbf{W}_i$, it becomes the patch embedding kernel for the corresponding spectral channel. Therefore, SSBD reduces the dynamic prediction target from $CM$ full embedding parameters to $CN$ wavelength-conditioned coefficients, where $N \ll M$. This compact subspace approximation is motivated by the observation that channel-specific embedding kernels share reusable spatial structures across bands and modalities. In practice, our ablation study in Fig.~\ref{fig:ablation_basis_number} shows that a basis number far smaller than the full kernel dimensionality is sufficient to achieve strong performance, indicating that this compact parameterization provides an effective trade-off between adaptation capacity and optimization stability.

\subsection{Spatial Basis Generation}

Effective spatial bases should retain the visual common sense learned by CLIP, since edges, textures, and local structures remain useful across heterogeneous sensors. Therefore, instead of learning the basis kernels from scratch, SSBD adopts a prior inheritance strategy. Given the original CLIP patch embedding weight $\mathbf{W}_{\mathrm{clip}}\in\mathbb{R}^{D\times 3\times K\times K}$, we first collapse the RGB channel dimension to extract a spectral-agnostic spatial prototype $\mathbf{B}_{\mathrm{base}}\in \mathbb{R}^{D\times K\times K}$: 
\begin{equation}
  \mathbf{B}_{\mathrm{base}}=\mathrm{AvgPool}(\mathbf{W}_{\mathrm{clip}}),
\end{equation}
where $D$ denotes the embedding dimension and $K$ is the patch size. This simple averaging operation removes the color-specific channel dependency while preserving the dominant spatial patterns encoded in the pretrained patch embedding. We then construct $N$ spatial bases by adding small-magnitude perturbations to the pretrained prototype:
\begin{equation}
  \mathbf{B}_{n}=\mathbf{B}_{\mathrm{base}}+\boldsymbol{\epsilon}_{n},
  \quad
  \boldsymbol{\epsilon}_{n}\sim\mathcal{N}(0,\sigma^2),
  \quad n=1,\ldots,N.
\end{equation}
In this way, all spatial bases are anchored in the pretrained feature space of the CLIP visual backbone at initialization, while the perturbations provide enough diversity for different bases to specialize during end-to-end multi-modal training.

\subsection{Wavelength-Conditioned Coefficient Generation}

A spectral band should not be treated as an isolated input channel. Its visual response depends on both its own spectral position and the wavelength configuration of the sensor, since neighboring bands, spectral gaps, and bandwidth density jointly determine what physical information the channel carries. Therefore, SSBD formulates coefficient generation as a sequence-level spectral reasoning problem: for each input sample, the model predicts how the shared spatial bases should be mixed according to the complete wavelength sequence.

Given the central wavelength $\lambda_i$ of the $i$-th channel, we first map it into a high-dimensional spectral token:
\begin{equation}
  \mathbf{e}_i = \phi\left(\mathrm{PE}(\lambda_i)\right),
\end{equation}
where $\mathrm{PE}(\cdot)$ denotes sinusoidal wavelength encoding and $\phi(\cdot)$ is a lightweight residual MLP. Next, to further capture inter-band dependencies, we prepend a learnable spectral query $\mathbf{e}_{\mathrm{s}}$ to the wavelength-token sequence and feed the full sequence into a Transformer encoder:
\begin{equation}
  [\mathbf{h}_s, \mathbf{h}_1,\ldots,\mathbf{h}_C]
  =
  \mathrm{Transformer}
  \left(
  [\mathbf{e}_{\mathrm{s}}, \mathbf{e}_1,\ldots,\mathbf{e}_C]
  \right),
\end{equation}
where $\mathbf{h}_s$ summarizes the global spectral context, and $\mathbf{h}_i$ is the contextualized representation of the $i$-th spectral band. Unlike an independent per-channel mapping, this self-attention formulation allows the coefficient of each band to depend on its relative location within the whole spectral profile. The basis mixing coefficients are then predicted by:
\begin{equation}
    \boldsymbol{\alpha}_i(\boldsymbol{\lambda})
    =
    \mathrm{Linear}_{\alpha}(\mathbf{h}_i),
    \quad
    \boldsymbol{\alpha}_i(\boldsymbol{\lambda})\in\mathbb{R}^{N}.
\end{equation}

In this way, $\boldsymbol{\alpha}_i$ becomes a wavelength-conditioned basis coordinate rather than a freely learned channel-specific parameter. The same spatial basis can be reused across sensors, while different wavelength configurations induce different basis mixtures for RGB, SAR, MSI, and HSI inputs. In particular, for SAR inputs, whose imaging mechanism differs from optical spectral sensing, the corresponding entry in $\boldsymbol{\lambda}$ denotes the C-band operating wavelength rather than an optical spectral band.

\subsection{Dynamic Spectral-Spatial Patch Embedding}
With the wavelength-conditioned coefficients, SSBD synthesizes a channel-wise patch embedding kernel for each input band by combining the shared spatial bases:
\begin{equation}
  \mathbf{W}_{i}
  =
  \sum_{n=1}^{N}\alpha_{i,n}(\boldsymbol{\lambda})\mathbf{B}_{n},
  \quad
  \mathbf{W}_{i}\in\mathbb{R}^{D\times K\times K}.
\end{equation}
Here, $\mathbf{B}_n$ is the reshaped kernel form of $\mathbf{b}_n$ in Eq.~\ref{eq:vectorized_basis}. In practice, to improve training stability, we further perform mean-variance normalization on the synthesized kernels, and denote the normalized kernel as $\widehat{\mathbf{W}}_i$. The normalized kernels are then used to embed an arbitrary-channel image by summing channel-wise patch responses:
\begin{equation}
  \mathbf{Z}
  =
  \sum_{i=1}^{C}
  \mathbf{X}_{i}\circledast \widehat{\mathbf{W}}_{i},
\end{equation}
where $\circledast$ denotes convolution operation. Thus, the output
$\mathbf{Z}\in\mathbb{R}^{D\times H'\times W'}$, where $H'$ and $W'$ denote the height and width of the embedded feature map, can be directly fed into the CLIP visual backbone. In this way, SSBD serves as a modality-adaptive embedding interface: spatial bases inherit the pretrained prior of CLIP, while wavelength-conditioned coefficients adapt these bases to arbitrary input channels without requiring a separate embedding layer for each sensor.

\subsection{Spectral-Context-Aware Mask-Based Alignment}
Text-conditioned feature selection~\cite{smartclip} is effective for fine-grained image-text alignment, as it can suppress irrelevant visual dimensions and emphasize caption-related semantics. We extend this method to heterogeneous remote sensing by introducing a spectral-context-aware Mask Network, which incorporates the spectral context produced by SSBD and makes feature selection aware of both caption semantics and sensor characteristics. Let $\mathbf{T}$ denote the text features from the text encoder, and $\mathbf{h}_s$ is the spectral token generated by SSBD. Subsequently, $\mathbf{T}$ and $\mathbf{h}_s$ are jointly input into the Mask Network $G_{\mathrm{mask}}(\cdot)$ to produce a text- and spectral-context-aware feature mask::
\begin{equation}
  \mathbf{m} = \sigma(G_{\mathrm{mask}}(\mathbf{T}, \mathbf{h}_s)),
\end{equation}
where $\sigma(\cdot)$ is the sigmoid activation function. We retain mask values larger than 0.5 to select the activated visual features. 
Following~\cite{smartclip}, the final training objective is:
\begin{equation}
  \mathcal{L} = \lambda_{\mathrm{align}}(\mathcal{L}_{\mathrm{sidm}} + \mathcal{L}_{\mathrm{dism}})
  +\lambda_{\mathrm{sparse}}\mathcal{L}_{\mathrm{sparse}},
\end{equation}
where $\mathcal{L}_{\mathrm{sidm}}$ encourages the mask to select visual dimensions that are most relevant to the paired text; $\mathcal{L}_{\mathrm{dism}}$ enforces the masked visual representation to remain discriminative across images; $\mathcal{L}_{\mathrm{sparse}}$ encourages compact feature activation and improves the efficiency of the selected representation. Further details are provided in the \textbf{supplementary material}.

\begin{table*}[t]
\centering
\renewcommand{\arraystretch}{1.1}
\setlength{\tabcolsep}{2.2pt}
\small
\begin{tabular}{ccccccccccc}
\hline
\multirow{2}{*}{Method} & \multirow{2}{*}{Backbone}
& \multicolumn{2}{c}{RGB} & \multicolumn{2}{c}{MSI}
& \multicolumn{2}{c}{HSI} & \multicolumn{2}{c}{SAR} & \multirow{2}{*}{Mean} \\
\cline{3-10}
& & I2T & T2I & I2T & T2I & I2T & T2I & I2T & T2I & \\
\hline
\multirow{2}{*}{LongCLIP$^{\dagger}$} & ViT-B/16 & 23.74/65.16 & 27.55/69.67 & 8.84/19.08 & 8.95/19.05 & 18.15/31.49 & 16.83/32.69 & 16.10/28.40 & 15.00/27.70 & 26.77 \\
& ViT-L/14 & 23.28/73.51 & \textbf{32.05/77.69} & 10.51/23.44 & 10.39/23.55 & 20.31/33.53 & 18.87/33.17 & 19.30/34.60 & \textbf{17.80}/33.20 & 30.32 \\
\hline
\multirow{2}{*}{HiMoCLIP$^{\dagger}$} & ViT-B/16 & 21.74/68.27 & 25.90/69.24 & 11.30/24.05 & 9.91/22.58 & 20.55/35.10 & 16.83/33.29 & 17.60/31.30 & 16.10/30.40 & 28.38 \\
& ViT-L/14 & 24.54/74.08 & 30.31/75.24 & 13.48/29.93 & 13.60/29.37 & 21.63/\textbf{40.62} & \textbf{22.36/42.67} & 18.40/33.00 & \textbf{18.10}/34.10 & 32.59 \\
\hline
\multirow{2}{*}{Ours} & ViT-B/16 & 30.90/75.79 & 30.36/76.79 & \textbf{14.38/31.46} & 13.58/\textbf{31.68} & \textbf{22.48}/38.70 & 20.91/38.34 & 19.20/35.00 & 17.10/34.20 & 33.18 \\
& ViT-L/14 & \textbf{31.45/76.71} & 31.41/77.40 & 14.35/30.47 & \textbf{14.16}/31.19 & 21.03/38.58 & 19.35/39.42 & \textbf{19.80/36.30} & 17.80/\textbf{37.10} & \textbf{33.53} \\
\hline
\end{tabular}
\caption{\textbf{Comparison of cross-modal retrieval performance on multi-source remote sensing data.} Each cell reports short-text/long-text Recall@1; Mean is the average of all Recall@1 values. $^{\dagger}$ denotes fine-tuned models.}
\label{tab:multimodel_results}
\end{table*}

\section{Dataset Construction}
To support the end-to-end training and evaluation of OmniRSCLIP, we construct OmniRS5M, a large-scale multi-modal remote sensing image-text dataset covering RGB, SAR, MSI, and HSI. OmniRS5M is built through a data synthesis pipeline that integrates multi-source data collection, modality-aware semantic distillation, structured XML organization, and dynamic text-supervised caption synthesis. During data collection, OmniRS5M is organized into train and test subsets: for source datasets with official public splits, we follow the released partitions whenever available; otherwise, we randomly divide the collected samples into disjoint subsets, ensuring image-level separation between training and testing. The final synthesized semantic files are further recombined into short and long caption candidates, providing multi-granularity supervision for contrastive pre-training. Detailed dataset sources, split protocols, prompt templates, XML schema, and synthesis rules are provided in the \textbf{supplementary material}.

Overall, OmniRS5M contains 4.67M remote sensing images and 73.90M text candidates, including 27.82M short captions and 46.08M long captions. It covers 4.32M RGB images, 236.13K SAR images, 71.87K MSI images, and 47.71K HSI images. To the best of our knowledge, OmniRS5M is the first large-scale remote sensing image-text dataset that jointly integrates RGB, SAR, MSI, and HSI modalities. 

\begin{table}[t]
\centering
\renewcommand{\arraystretch}{1.1}
\setlength{\tabcolsep}{4.2pt}
\small
\resizebox{\columnwidth}{!}{
\begin{tabular}{llcccc}
\hline
Method & Backbone & \multicolumn{2}{c}{Image-to-Text} & \multicolumn{2}{c}{Text-to-Image} \\
\cline{3-6}
& & R@1 & R@5 & R@1 & R@5 \\
\hline
\noalign{\vskip 1.5pt}
\rowcolor{gray!5}
\multicolumn{6}{c}{\rule{0pt}{2ex}\textbf{\textit{Results on RSITMD Dataset}}} \\
\noalign{\vskip 1.5pt}
\hline
\multirow{2}{*}{CLIP}       & ViT-B/16 & 7.96  & 21.46 & 8.36  & 26.11 \\
                            & ViT-L/14 & 10.62 & 27.43 & 10.04 & 33.33 \\
\cline{2-6}
\multirow{2}{*}{DGTRSCLIP}  & ViT-B/16 & 21.90 & 39.82 & 15.18 & 40.80 \\
                            & ViT-L/14 & 22.12 & 42.26 & 17.35 & 43.27 \\
\cline{2-6}
\multirow{2}{*}{RemoteCLIP} & ViT-B/32 & 27.88 & 50.66 & 22.17 & 56.46 \\
                            & ViT-L/14 & 28.76 & 52.43 & \textbf{23.63} & \textbf{59.42} \\
\cline{2-6}
\multirow{2}{*}{Ours}       & ViT-B/16 & \textbf{30.53} & \textbf{53.54} & 23.10 & 51.37 \\
                            & ViT-L/14 & \textbf{33.19} & \textbf{56.19} & 22.61 & 55.44 \\
\hline
\noalign{\vskip 1.5pt}
\rowcolor{gray!5}
\multicolumn{6}{c}{\rule{0pt}{2ex}\textbf{\textit{Results on RSICD Dataset}}} \\
\noalign{\vskip 1.5pt}
\hline
\multirow{2}{*}{CLIP}       & ViT-B/16 & 5.67  & 14.82 & 5.29  & 16.96 \\
                            & ViT-L/14 & 6.59  & 17.75 & 4.96  & 18.72 \\
\cline{2-6}
\multirow{2}{*}{DGTRSCLIP}  & ViT-B/16 & 11.53 & 29.19 & 9.92  & 28.60 \\
                            & ViT-L/14 & 14.36 & 28.36 & 11.22 & 31.86 \\
\cline{2-6}
\multirow{2}{*}{RemoteCLIP} & ViT-B/32 & 17.02 & 37.97 & 13.71 & 37.11 \\
                            & ViT-L/14 & 18.39 & 37.42 & 14.73 & 39.93 \\
\cline{2-6}
\multirow{2}{*}{Ours}       & ViT-B/16 & 23.06 & 45.11 & 17.00 & 42.34 \\

                            & ViT-L/14 & \textbf{23.97} & \textbf{46.57} & \textbf{17.75} & \textbf{44.87} \\
\hline
\end{tabular}
}
\caption{Comparison of cross-modal retrieval performance on RSITMD and RSICD.}
\label{tab:retrieval_results}
\end{table}

\begin{figure}[t]
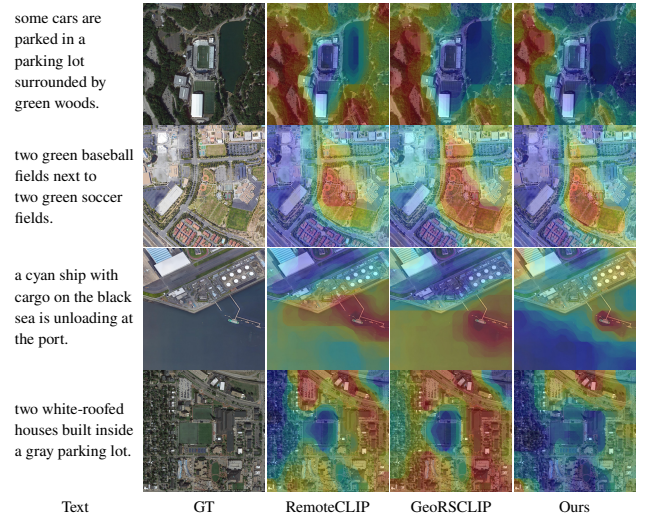

\centering
{
\setlength{\tabcolsep}{\colspacing}
\renewcommand{\arraystretch}{\arraystretchval}
\newlength{\textcolwidth}
\setlength{\textcolwidth}{0.19\columnwidth}
\newlength{\imgcolwidth}
\setlength{\imgcolwidth}{0.19\columnwidth}
\newlength{\imgsize}
\setlength{\imgsize}{0.19\columnwidth}
\begin{adjustbox}{max totalsize={\columnwidth}{0.95\textheight},center}
\begin{tabular}{
  >{\centering\arraybackslash}p{\textcolwidth}
  @{\hspace{\textcolgap}}
  *{4}{>{\centering\arraybackslash}p{\imgcolwidth}}
}
  \textcell{some cars are parked in a parking lot surrounded by green woods.} & \imgsubfig[2]{img/intr.pdf} & \imgsubfig[3]{img/intr.pdf} & \imgsubfig[4]{img/intr.pdf} & \imgsubfig[5]{img/intr.pdf} \\[\rowspacing]
  \textcell{two green baseball fields next to two green soccer fields.} & \imgsubfig[6]{img/intr.pdf} & \imgsubfig[7]{img/intr.pdf} & \imgsubfig[8]{img/intr.pdf} & \imgsubfig[9]{img/intr.pdf} \\[\rowspacing]
  \textcell{a cyan ship with cargo on the black sea is unloading at the port.} & \imgsubfig[10]{img/intr.pdf} & \imgsubfig[11]{img/intr.pdf} & \imgsubfig[12]{img/intr.pdf} & \imgsubfig[13]{img/intr.pdf} \\[\rowspacing]
  \textcell{two white-roofed houses built inside a gray parking lot.} & \imgsubfig[14]{img/intr.pdf} & \imgsubfig[15]{img/intr.pdf} & \imgsubfig[16]{img/intr.pdf} & \imgsubfig[17]{img/intr.pdf} \\[\headrowspacing]
  \headcell{Text} & \headcell{GT} & \headcell{RemoteCLIP} & \headcell{GeoRSCLIP} & \headcell{Ours} \\
\end{tabular}
\end{adjustbox}
}
\caption{Aggregated heatmaps of different methods.}
\label{fig:selo}
\end{figure}

\section{Experiment}
\subsection{Implementation Details}
We implement OmniRSCLIP with CLIP-style vision backbones, including ViT-B/16 and ViT-L/14. For both backbones, the original fixed-channel patch embedding layer is replaced by the proposed SSBD module, and the number of spatial bases is set to 64. The loss weights $\lambda_{\mathrm{align}}$ and $\lambda_{\mathrm{sparse}}$ are set to 10 and 2, respectively. To mitigate training data imbalance, we adopt a two-stage training strategy, training each stage for 5 epochs. The first stage uses only RGB image-text pairs, while the second incorporates multi-modal data. The global batch sizes are 1024 and 384 in the first and second stages, respectively. For both the vision and text encoders, the learning rate is $1\times10^{-4}$ in the first stage and $1\times10^{-5}$ in the second, while that of the Mask Network remains $1\times10^{-3}$ throughout. All ViT-B/16 experiments use 8 NVIDIA A100 GPUs, whereas ViT-L/14 experiments use 16.

\begin{table}[t]
\centering
\renewcommand{\arraystretch}{1.1}
\small
\begin{tabular}{llcccc}
\hline
\multirow{2}{*}{Backbone} & \multirow{2}{*}{Method} & \multicolumn{4}{c}{Testing Dataset} \\
\cline{3-6}
& & AID & WHU & OPT & RSC \\
\hline
\multirow{3}{*}{ViT-B/16}
& CLIP & 66.38 & 81.29 & 73.44 & 61.44 \\
& DGTRSCLIP & 74.97 & 90.15 & 85.43 & 73.78 \\
\rowcolor{gray!10}
& Ours & \textbf{85.21} & \textbf{95.32} & \textbf{89.68} & \textbf{82.14} \\
\hline
\multirow{5}{*}{ViT-L/14}
& CLIP & 69.47 & 85.77 & 80.86 & 63.67 \\
& GeoRSCLIP & 75.46 & 92.14 & 85.91 & 75.89 \\
& RemoteCLIP & \textbf{88.53} & 96.12 & 88.17 & 70.78 \\
& DGTRSCLIP & 77.49 & 93.03 & 89.19 & \textbf{82.63} \\
\rowcolor{gray!10}
& Ours & 85.35 & \textbf{96.52} & \textbf{90.32} & 79.30 \\
\hline
\end{tabular}
\caption{\textbf{Comparison of zero-shot classification accuracy.} AID, WHU, OPT, and RSC denote AID~\cite{aid}, WHU-RS19~\cite{whurs19}, OPTIMAL-31~\cite{opt}, and RS-C11~\cite{rsc11}, respectively.}
\label{tab:zeroshot_results}
\end{table}

\subsection{Cross-Modal Retrieval}
Following~\cite{remoteclip, dgtrsclip}, we adopt cross-modal retrieval as the primary task to evaluate the performance of OmniRSCLIP. Specifically, in contrast to existing methods that predominantly focus on the RGB modality, we extend the evaluation scope to diverse remote sensing modalities: SAR, MSI, and HSI. 

First, we evaluate OmniRSCLIP on public RGB short-text benchmarks, as summarized in Tab.~\ref{tab:retrieval_results}. Although designed for heterogeneous sensors beyond RGB, OmniRSCLIP remains competitive on standard RGB retrieval tasks. With the ViT-L/14 backbone, it achieves the best image-to-text results on RSITMD~\cite{rsitmd} and the best performance across all reported metrics on RSICD~\cite{rsicd}. These results show that OmniRSCLIP preserves strong RGB-domain alignment while extending CLIP to arbitrary-modality remote sensing data.

Beyond standard RGB benchmarks, Tab.~\ref{tab:multimodel_results} evaluates multi-source image-text retrieval on RGB, SAR, MSI, and HSI. Since existing RGB-oriented VLMs cannot directly handle arbitrary-channel inputs, we replace the patch embedding layers of LongCLIP~\cite{longclip} and HiMoCLIP~\cite{himoclip} with SSBD and fine-tune them under the same multi-modal setting. Thus, this comparison focuses on different baselines under a unified arbitrary-channel input interface, while different input adaptation strategies are evaluated in Tab.~\ref{tab:ablation_clip_init}. We report both short- and long-text retrieval to examine the effect of semantic richness. OmniRSCLIP achieves the best mean performance, suggesting its effectiveness for unified alignment across heterogeneous modalities.

\subsection{Zero-shot Classification}
Zero-shot classification evaluates the category-level transferability of the learned vision-language space without task-specific supervision. We average prompt embeddings into class prototypes and predict the class with the highest image-text cosine similarity. As shown in Tab.~\ref{tab:zeroshot_results}, OmniRSCLIP shows strong classification performance: ViT-B/16 achieves the best results among the compared methods on all four datasets, while ViT-L/14 also remains consistently competitive. These results indicate that OmniRSCLIP preserves category-level recognition ability while extending CLIP to heterogeneous sensor inputs.

\subsection{Semantic Localization}
To further examine fine-grained cross-modal grounding, we provide qualitative semantic localization results on the AIR-SLT~\cite{yuan2022learning} dataset. Given a natural-language query, we decompose each image into multi-scale local views, compute their similarities with the query, and project the responses back to the original image to form a heatmap. As shown in Fig.~\ref{fig:selo}, compared with existing remote sensing VLMs, OmniRSCLIP produces more concentrated responses on text-related regions and fewer activations on irrelevant areas, indicating stronger fine-grained grounding in complex remote sensing scenes.

\begin{table}[t]
\centering
\small
\renewcommand{\arraystretch}{1.08}
\resizebox{\columnwidth}{!}{
\begin{tabular}{cccccc}
\hline
SSBD & Mask Network & RGB & SAR & MSI & HSI \\
\hline
 &  & 65.70 & 47.67 & 34.89 & 43.87 \\
$\checkmark$ &  & 63.82 & 47.31 & 38.16 & 48.94 \\
 & $\checkmark$ & 67.92 & 47.29 & 35.94 & 52.48 \\
\rowcolor{gray!10}
$\checkmark$ & $\checkmark$ & \textbf{69.08} & \textbf{48.81} & \textbf{41.10} & \textbf{53.64} \\
\hline
\end{tabular}
}
\caption{\textbf{Ablation of core components.}}
\label{tab:ablation_components}
\end{table}

\begin{table}[t]
\centering
\small
\renewcommand{\arraystretch}{1.08}
\setlength{\tabcolsep}{6pt}
\resizebox{\columnwidth}{!}{
\begin{tabular}{lccccc}
\hline
Method & CLIP Prior & RGB & SAR & MSI & HSI \\
\hline
PCA & $\times$ & 65.70 & 47.67 & 34.89 & 43.87 \\
PCA & $\checkmark$ & 67.92 & 47.29 & 35.94 & 52.48 \\
DOFA & $\times$ & 67.54 & 46.94 & 40.95 & 52.31 \\
SSBD & $\times$ & 64.92 & 45.69 & 36.67 & 49.10 \\
\rowcolor{gray!10}
SSBD & $\checkmark$ & \textbf{69.08} & \textbf{48.81} & \textbf{41.10} & \textbf{53.64} \\
\hline
\end{tabular}
}
\caption{\textbf{Effect of preserving the CLIP patch-embedding prior.} Results are per-modality average recall.}
\label{tab:ablation_clip_init}
\end{table}

\begin{figure}[t]
\centering
\subfloat[Image-to-text]{
  \includegraphics[width=0.475\columnwidth]{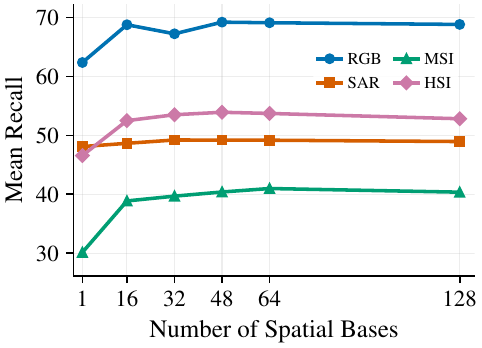}
}
\hfill
\subfloat[Text-to-image]{
  \includegraphics[width=0.475\columnwidth]{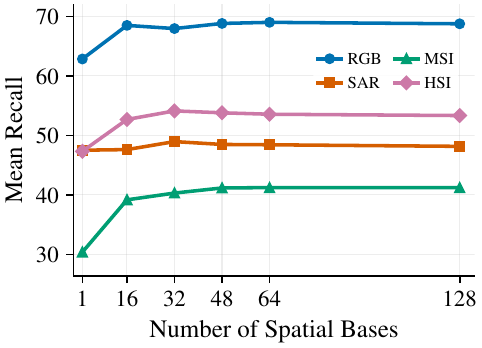}
}
\caption{Effect of the number of spatial bases in SSBD.}
\label{fig:ablation_basis_number}
\end{figure}

\subsection{Ablation Study}
We conduct ablation studies using ViT-B/16 and, unless otherwise specified, report the per-modality mean of R@1, R@5, and R@10 over both short- and long-text retrieval.

\subsubsection{Effect of Core Components}
Table~\ref{tab:ablation_components} validates the contributions of SSBD and Mask Network. Without SSBD, MSI and HSI inputs are projected to three channels by PCA, while SAR inputs are replicated to fit the original CLIP interface. In the absence of Mask Network, SSBD improves the average recall on MSI and HSI from 34.89/43.87 to 38.16/48.94, demonstrating its ability to better exploit spectral information than fixed three-channel adaptation. Incorporating Mask Network further improves performance across all modalities, with the full configuration achieving the best overall results. This confirms their complementarity: SSBD provides modality-adaptive spectral embedding, while the Mask Network further selects alignment-relevant features using semantic and spectral context.

\subsubsection{Effect of CLIP Prior}
We further examine the effect of preserving the CLIP visual prior, defined as initializing patch embeddings or spatial bases from vanilla CLIP rather than random weights. As shown in Tab.~\ref{tab:ablation_clip_init}, prior inheritance consistently improves the overall performance of both PCA- and SSBD-based adaptation. In particular, CLIP-initialized SSBD achieves gains across all four modalities and outperforms DOFA, demonstrating that basis decomposition effectively adapts heterogeneous inputs while retaining transferable visual knowledge from CLIP.

\subsubsection{Effect of the Number of Spatial Bases}
Unlike full-kernel generation, SSBD only learns a compact basis set and predicts wavelength-conditioned coefficients for basis recombination. As shown in Fig.~\ref{fig:ablation_basis_number}, using only one basis leads to a clear performance drop, because it is equivalent to applying the same spatial transformation to all spectral bands and therefore cannot model inter-band spectral differences. Increasing the basis number quickly improves the performance on MSI and HSI, while the curves become stable around 48 to 64 bases. Further increasing the dictionary to 128 brings limited additional gains. We thus adopt 64 bases by default for a favorable balance between performance and complexity.

\section{Conclusion}
This paper introduces OmniRSCLIP, a unified vision-language framework for heterogeneous remote sensing image-text alignment. With SSBD as a modality-adaptive embedding interface and spectral-text-aware feature selection, OmniRSCLIP enables a single CLIP-based model to align RGB, SAR, MSI, and HSI observations within a shared semantic space. Experiments on retrieval, zero-shot classification, qualitative semantic localization, and ablations show that OmniRSCLIP extends CLIP beyond RGB imagery while preserving strong semantic transferability.

\bibliography{main}


\end{document}